\documentclass[10pt,conference]{IEEEtran}
\pdfoutput=1

\IEEEoverridecommandlockouts
\usepackage{xspace}
\usepackage{pifont}
\usepackage{pgfplots}
\usepackage{subcaption}
\usepackage{array}
\usepackage{svg}

\usepackage[normalem]{ulem}
\usepackage[linesnumbered,ruled,vlined]{algorithm2e}
\usepackage{amsmath}

\SetCommentSty{mycommfont}
\usepackage{cite}
\usepackage{multirow}
\usepackage{hyperref}

\usepgfplotslibrary{
  groupplots,
units, }

\pgfplotsset{compat=1.3}
\usepackage[binary-units=true]{siunitx}
\usepackage{balance}
\usepackage{hyperref}

\usepackage{amsmath,amssymb,amsfonts}
\usepackage{graphicx}
\usepackage{textcomp}
\usepackage{xcolor}
\usepackage{tabularx}
\usepackage{booktabs}
\usepackage{luatex85} % Sometimes necessary for compatibility

\newcommand{\ie}{\emph{i.e.,}\xspace}
\newcommand{\eg}{\emph{e.g.,}\xspace}
\newcommand{\wrt}{\emph{w.r.t.}\xspace}

\newcommand{\globalModel}{$W_{global}^{t+1}$}

\newcommand{\tool}{{TraceFL}\xspace}

\usepackage{fancyhdr}
\usepackage{xcolor} % For coloring the header text

\def\BibTeX{{\rm B\kern-.05em{\sc i\kern-.025em b}\kern-.08em
T\kern-.1667em\lower.7ex\hbox{E}\kern-.125emX}}
\makeatletter
\def\ps@IEEEtitlepagestyle{%
    \fancyhf{} % Clear all header and footer fields
    \fancyhead[C]{\textcolor{blue}{\textbf{Accepted at 2025 IEEE/ACM 47th International Conference on Software Engineering (ICSE)}}} % Centered header
    \fancyfoot[C]{\thepage} % Centered footer with page number
    \renewcommand{\headrulewidth}{0pt} % Remove header line
    \renewcommand{\footrulewidth}{0pt} % Remove footer line
}
\makeatother
\begin{document}

\title{TraceFL: Interpretability-Driven Debugging in Federated Learning via Neuron Provenance}

\author{\IEEEauthorblockN{Waris Gill}
  \IEEEauthorblockA{\textit{Computer Science Department} \\
    \textit{Virginia Tech}\\
    Blacksburg, USA \\
  waris@vt.edu}
  \and
  \IEEEauthorblockN{Ali Anwar}
  \IEEEauthorblockA{\textit{Computer Science and Engineering Department} \\
    \textit{University of Minnesota}\\
    Minneapolis, USA \\
  aanwar@umn.edu}
  \and
  \IEEEauthorblockN{Muhammad Ali Gulzar}
  \IEEEauthorblockA{\textit{Computer Science Department} \\
    \textit{Virginia Tech}\\
    Blacksburg, USA \\
  gulzar@cs.vt.edu}
  \and

}

\maketitle

% Set up headers for subsequent pages
\pagestyle{fancy}
\fancyhf{} % Clear all header and footer fields
\fancyhead[C]{\textcolor{blue}{\textbf{Accepted at 2025 IEEE/ACM 47th International Conference on Software Engineering (ICSE)
}}} % Centered header
\fancyfoot[C]{\thepage} % Centered footer with page number
\renewcommand{\headrulewidth}{0pt} % Remove header line
\renewcommand{\footrulewidth}{0pt} % Remove footer line

\begin{abstract}
  In Federated Learning, clients train models on local data and send updates to a central server, which aggregates them into a global model using a fusion algorithm. This collaborative yet privacy-preserving training comes at a cost. FL developers face significant challenges in attributing global model predictions to specific clients. Localizing responsible clients is a crucial step towards (a) excluding clients primarily responsible for incorrect predictions and (b) encouraging clients who contributed high-quality models to continue participating in the future. Existing ML debugging approaches are inherently inapplicable as they are designed for single-model, centralized training.

We introduce {\tool}, a fine-grained {\em neuron provenance} capturing mechanism that identifies clients responsible for a global model's prediction by tracking the flow of information from individual clients to the global model. Since inference on different inputs activates a different set of neurons of the global model, \tool dynamically quantifies the significance of the global model's neurons in a given prediction, identifying the most crucial neurons in the global model. It then maps them to the corresponding neurons in every participating client to determine each client's contribution, ultimately localizing the responsible client.
We evaluate \tool on six datasets, including two real-world medical imaging datasets and four neural networks, including advanced models such as GPT. \tool achieves 99\% accuracy in localizing the responsible client in FL tasks spanning both image and text classification tasks. At a time when state-of-the-art ML debugging approaches are mostly domain-specific (\eg image classification only), {\tool} is the first technique to enable highly accurate automated reasoning across a wide range of FL applications.

\end{abstract}


\section{Introduction}

Federated Learning (FL) offers distributed training that enables multiple clients to collaboratively train a global model without sharing raw data~\cite{mcmahan2017communication, jiang2020federated, rieke2020future, long2020federated, zheng2021applications}. In a typical FL setup, individual clients, such as healthcare institutions, train models on their local data. These local models are then aggregated on a central server to form a comprehensive global model, all without transferring sensitive client data. The resulting global model, a fusion of all clients' models, is then used in production to make predictions on unseen data.

The complexity of FL systems, however, introduces unique debugging challenges. When a global model makes a prediction, whether correct or incorrect, a key question arises: \emph{which client(s) is primarily responsible for a global model's output?} This question is akin to debugging software, where understanding the impact of each input and the line of code on the software's output is crucial. Addressing this debugging question is vital for the effective deployment, maintenance, and accountability of FL applications. For example, FL developers face challenges in identifying and rewarding clients responsible for successful classifications. This recognition is crucial to encourage their continued participation in future incentivized FL rounds~\cite{cho2022federate}. There is mature evidence that such practice significantly improves the FL model's quality~\cite{zhan2021survey}. Similar debugging is key in localizing {\em faulty clients} that may transfer an inaccurate model for aggregation, which can result in a dangerously low-quality global model ~\cite{ bagdasaryan2020backdoor, bhagoji2019analyzing,gill2023feddefender, xhemrishi2023fedgt, ali2023survey}.

\noindent{\underline{\textit{\textbf{Problem.}}}} {\em In federated learning, the client(s) most responsible for a global model's prediction are the ones trained on data that contains the predicted labels}. This is analogous to finding influential training samples in classical machine learning~\cite{koh2017understanding}. However, the two domains, single model-based centralized ML and FL, are fundamentally different. Existing influence-based debugging approaches in ML~\cite{stokes2021preventing, 8473440, baracaldo2017mitigating, rupprecht2020improving} and regular software~\cite{10.14778/2095686.2095693, 10.14778/3402755.3402768, akoush2013hadoopprov} require transparent access to data including all data manipulation operations applied on the input data. When applied to FL, these approaches will require end-to-end monitoring of clients' training (\ie require access to clients' data), which is prohibited in FL. More broadly, ML influence and interpretability-based debugging approaches target a single model in which the debugging is restricted to identifying the training data. In contrast, debugging in FL entails isolating a client's model among many. This paper addresses the following debugging problem in FL: {\em Given the global model inference on an input in FL, how can we identify the client(s) most responsible for the inference?}

\noindent{\underline{\textit{\textbf{Challenges.}}}} Determining a client's influence on the global model is challenging. Clients are randomly sampled in each round, each possessing unique data and contributing differently to the global model. Thus, the influence of a client on the global model is dynamic, non-uniform, and changes across rounds, making it difficult to link the global model's behavior to a specific client. The FL protocol restricts access to client-side training, turning FL configuration into a nearly black-box setting. Additionally, clients' models are collections of neuron weights that are individually uninterpretable. Static analysis of models' weights to measure clients' influence is ineffective because clients' models are intrinsically different in terms of weights. Furthermore, neural networks today comprise millions of neurons (\eg GPT-3 has 175 billion parameters~\cite{OPTQ}). Considering all neurons equally, in such cases, would lead to imprecise and incorrect debugging.

FL is increasingly used for domains other than vision using various neural networks, such as transformer and convolutional neural networks (CNNs). Designing a generic FL debugging approach is a major challenge. For instance, transformers contain a self-attention mechanism that allows the model to focus on different parts of the input sequence. This mechanism is usually not seen in CNNs; instead, it uses a convolutional layer to detect the special patterns in the input data. Additionally, these architectures use different activation functions such as Rectified Linear Unit (ReLU)~\cite{10.5555/3104322.3104425} and Gaussian Error Linear Unit (GELU)~\cite{hendrycks2016gaussian}, introducing another source of complications.

\noindent{\underline{\textit{\textbf{Our Contribution.}}}} We present the concepts of \textbf{{\em neuron provenance}}, a fine-grained lineage-capturing mechanism that formulates the flow of information in the fusion algorithm from multiple clients' models into a global FL model, ultimately influencing the predictions of the global model. Using neuron provenance, we determine the precise magnitude of contributions of participating clients towards the global model's prediction. We materialize the idea of neuron provenance in \tool, which runs at the aggregator (\ie central server) and requires no instrumentation on the client side.

\tool is designed with the following insights. Since a global model consists of millions of neurons, we observe that a dynamic subset of neurons activates in response to a given input, and not all neurons contribute equally to a prediction~\cite{olah2018building, yu2022spatl}.
Using this insight, \tool quantifies the contribution of these neurons in the global model's prediction by computing the gradient of the neurons \wrt to the prediction. Such neuron-level gradients reveal the neuron's output impact on the global model's prediction and thus reduce the scope of important neurons.

\tool then maps the global model's important neurons to the corresponding neurons in each client's model and computes the contribution of each client's neuron to the corresponding global model's neuron. At this stage, \tool computes the end-to-end \textbf{neuron provenance} of the global model prediction with the magnitude of the contribution of each client's neurons.
Finally, \tool aggregates the contributions of each client. The client with the highest contribution is deemed the most responsible for the given prediction.

\noindent{\underline{\textit{\textbf{Evaluations.}}}} We demonstrate \tool's effectiveness, generalizability, and robustness by evaluating its client localization accuracy on both image and language models under various commercial data distributions and differential privacy methods.  We evaluate \tool on four state-of-the-art neural networks: ResNet~\cite{he2015deep}, DenseNet~\cite{huang2017densely}, BERT~\cite{Devlin2019BERTPO}, and GPT~\cite{radford2018improving} and using six datasets including two real-world medical imaging datasets~\cite{kather2019predicting, xu2019efficient, bilic2023liver}.
\tool achieves an average accuracy of 99\% in localizing the responsible client across 30 unique FL settings, spanning both correct and misprediction scenarios. For fault localization in FL, \tool achieves an average accuracy of 99\%, compared to 32\% by the existing technique~\cite{feddebug}, demonstrating \tool's effectiveness in real-world FL deployments. Additionally, we test \tool's robustness against varying data distributions and differential privacy settings and find that \tool remains robust and effective. We also vary the number of clients, increasing it up to 1000, and find that \tool is both scalable and efficient. Overall, we evaluate \tool on  20,600 trained client models. These experiments exceed prior research's evaluation complexity and fully represent commercial FL usage~\cite{konečný2018federated, Avdiukhin2021FederatedLU, bonawitz2019towards, li2022federated, wang2020tackling}. \tool is implemented in Flower FL~\cite{beutel2020flower} and compatible with GPU for parallel processing of \textbf{neuron provenance} for compute-intensive models (\eg GPT).

\tool advances the state of FL debugging with the following core contributions:

\begin{itemize}
  \item \tool localizes the responsible clients for a given prediction without modifying the underlying fusion algorithm. Moreover, it does not require access to clients' training and can solely determine clients' contributions at the central aggregator.

  \item \tool introduces a unique concept of \textbf{{\em neuron provenance}} for FL applications to capture the dynamic contribution of each client, which helps rank clients based on the contribution to a given prediction. \tool efficiently tracks the contribution of clients in large models like GPT containing millions of parameters.

  \item \tool achieves 99\% localization accuracy in localizing the responsible client in FL. \tool's localization accuracy remains high during localizing a faulty client where existing baseline\cite{feddebug} achieves 32\%.

  \item \tool is the first approach that is equally effective on transformers and CNNs. Even the most sophisticated ML debugging approaches work on single model architecture and data domains, \ie either CNNs or transformers.

  \item \tool significantly advances the field of debugging and interpretability in FL, addressing open challenges in FL~\cite{feddebug, kairouz2021advances} and work with differential privacy and real-world data distributions among FL clients.
\end{itemize}

\noindent{\underline{\textit{\textbf{Source Code.}}}} \tool's artifact is available at \textcolor{blue}{ \url{https://github.com/SEED-VT/TraceFL}}.

\begin{figure*}[t]
  \centering
  \includegraphics[width=0.90\textwidth]{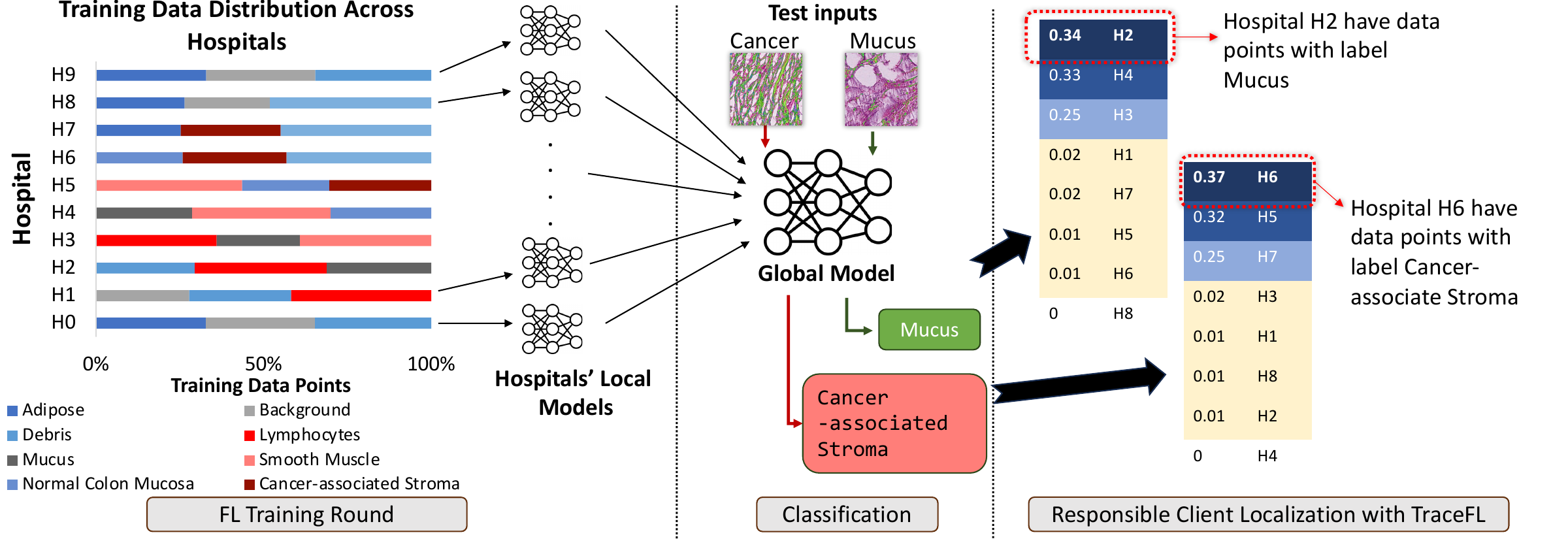}
  \caption{Illustration of training, testing, and localization phases of the real-world motivating example. The FL global model correctly classifies two colon pathology images (original labels `Cacner-associated Stroma' and `Mucus'). During responsible client localization, \tool accurately identifies the client most responsible for the prediction, i.e., clients trained on data points with labels Mucus (Hospital H2) and `Cancer-associated Stroma' (Hospital H6).}
  \label{fig:motivating-example}
  \vspace{-3ex}
\end{figure*}

\section{Background and Motivation}
\subsection{Federated Learning}
Federated Learning enables multiple clients (\eg mobile devices, organizations) to train a shared model without sharing their data. This allows the model to be trained using distributed data, which can be useful in cases where data is distributed across multiple devices or organizations and cannot be easily collected and centralized. One algorithm of FL is Federated Averaging (FedAvg)~\cite{mcmahan2017communication}, which uses the following equation to update the global model at each round of the training process:

\vspace{-3ex}
\begin{equation} \label{eq:1}
  W_{global}^{t+1} =  \sum_{k=1}^K \frac{n_k}{n} W_k^{(t)}
\end{equation}
% \vspace{-2ex}

where $W_k^{(t)}$ and $n_k$ represent received weights and size of training data of client $k$ in each round $t$, respectively. The variable $n$ represents the total number of data points from all clients, and it is calculated as $n = \sum_{k=1}^{K} n_k$. The equation states that the global model $W^{t+1}$ at the next round is the average of the local models from all participating clients at the current round. In each round, the clients first train their local models using their own data, then send the parameters (\eg $W_k^{(t)}$, $n_k$) to the central server. The central server averages the model parameters to produce a global model, which is then sent back to the participating devices. This process is repeated for multiple rounds (\eg $t$ from 1 to 100), with each client updating its local model using the global model from the previous round. The final global model is the result of the federated averaging process.

FL has variations such as Vertical FL~\cite{liu2019communication} and Personalized FL~\cite{t2020personalized}. \tool primarily focuses on horizontal FL~\cite{mcmahan2017communication}, similar to previous work on fault localization in FL~\cite{feddebug, gill2023feddefender}. A typical FL setup involves a few to thousands of clients, such as mobile devices, healthcare institutions, or enterprises. FL clients exhibit diversity in data distribution and computational resources. Data is often non-independently and identically
distributed (Non-IID), with varying sizes across clients, such as hospitals specializing in different medical conditions (Figure~\ref{fig:motivating-example}). All participating clients use the same neural network architecture, ensuring compatibility during model aggregation. Additionally, clients have heterogeneous hardware and network capabilities (e.g., smartphones to powerful servers), impacting their participation and training consistency~\cite{khan2024float}.

\subsection{Motivation}
 Suppose a developer deploys an FL system to diagnose colon diseases based on colon pathology images, as shown in Figure~\ref{fig:motivating-example}. In this FL system,
ten hospitals, identified as H0 to H9, collaborate to train a global FL model.  Each hospital trains its local model, which is then aggregated to form a global model. The classification stage in Figure~\ref{fig:motivating-example} shows a scenario where the global model makes a correct prediction on new test colon pathology images (\eg `Cancer-associated Stroma' and `Mucus'). Since these are correct predictions, the FL developer aims to determine which hospital is most responsible for these correct predictions so that it can be encouraged to participate in future rounds. Since the training data is protected under the privacy of medical records and inaccessible to the developer, it is challenging to identify the responsible hospital by inspecting the raw model weights shared by the hospitals.

To address this issue, the developer decides to use \tool to identify the most responsible client behind the correct predictions. When enabled during the global model's prediction, \tool localizes the hospital with H2 as the one responsible for the prediction of `Mucus', and the hospital H6 as the one responsible for the prediction of `Cancer-associated Stroma'. More specifically, as shown on the right in Figure~\ref{fig:motivating-example}, \tool ranks hospitals (\ie clients) based on their contributions to each prediction. The score associated with each hospital quantifies how responsible a hospital is for that prediction. The training data distribution on the left shows that H2's training data include data points labeled `Mucus', whereas H6's training data include data points labeled `Cancer-associated Stroma'.

Conversely, hospital H1's contribution is 0.02 and 0.01 in the two predictions because it does not have any data points with the labels `Mucus' or `Cancer-associated Stroma', respectively. Detailed evaluations on two real-world medical imaging datasets are presented in Section~\ref{sec:correct-predictions-localization}. Localizing the clients responsible for the prediction with \tool serves as a basis for designing advanced incentivization approaches.

\section{{Challenges in Debugging FL}}
\label{sec:challenges}

Federated Learning poses several challenges in designing a debugging technique that reasons about a global model's prediction on an input. Unlike traditional ML training, where training data can be easily analyzed, the FL global model (\globalModel) is not directly trained on the data. Instead, the global model is generated by fusing clients' models together across many rounds using popular fusion algorithms. With no insight into the training, it is challenging to identify how different clients influence the global model's behavior.

State-of-the-art neural networks, such as Transformers (BERT, GPT) and CNNs (ResNet, DenseNet), have varying structures, activation functions, and numbers of parameters. For example, GPT~\cite{radford2018improving} is a 37-layer (12 block) Transformer architecture with GELU~\cite{hendrycks2016gaussian} activation function and 117 million parameters.
DenseNet~\cite{huang2017densely} has 121 layers, 8 million parameters, and uses the ReLU~\cite{10.5555/3104322.3104425} activation function. The fusion algorithm or the global model does not inherently provide any information about individual clients' contributions. Thus, tracking a client's contribution among millions of parameters is a significantly challenging task.

Moreover, the data distribution across FL training rounds is non-identical; clients rarely have the same data point. Not all clients participate in every round, and some clients may contain only a few data points to train their local model. The class label distribution also varies across clients. Such variability causes more hurdles in precisely reasoning about a global model's behavior. Simply tracking the static weights of the global model is inadequate, as different sets of neurons are activated on different inputs, and not all neurons have equal importance. Inspecting individual clients' models does not help understand the client's contribution to the global model prediction, as it will not capture the cumulative behavior of the global model.

\begin{algorithm}[t]

  {\footnotesize
    \SetKwInput{KwInput}{Input}                % Set the Input

    \SetKwInput{KwOutput}{Output}
    \KwInput{Let $clients$ be the list of clients' models participating in the FL training round.}
    \KwInput{Let $global\_model$ be the aggregated global model of $clients$ models after the end of a training round}
    \KwInput{Let $test\_input$ be an input}
    \KwOutput{$client2norm\_contribution$ contains the contribution of each client in the prediction of $test\_input$}

    % \SetKwProg{Fn}{Function}{:}{}
    \SetKwProg{Fn}{Function}{:}{}
    {

      $\text{activated\_neurons} = [\ ]$\;

      \tcp{Section~\ref{sec:Measuring Neuron's Contribution} (Equation~\ref{eq:neuron-contribution})}
      $y = \text{global\_model}(\text{test\_input})$\;
      \For{each $\text{neuron}$ \textbf{in} $\text{global\_model}$}{
        $\text{activated\_neurons}.\text{append}(\text{neuron})$\;
      }

      $\text{neuron2grad} = y.\text{backward()}$\;

      \tcp{Section~\ref{sec:Neuron-Provenance-Across-Fusion} }
      $\text{neuron2prov} = \{\}$\;
      \For{$\text{neuron}$ \textbf{in} $\text{activated\_neurons}$}{
        $\text{neuron2prov}[\text{neuron}] = \{\}$\;
        \For{$\text{client}$ \textbf{in} $\text{clients}$}{
          % $\text{client\_contribution} = \text{\CalculateClientContribution}(\text{neuron}, \text{client})$\;
          $\text{cont} = 0$\;
          \For{$\text{feature}$ \textbf{in} $\text{neuron.input\_features}$}{
            $\text{cont} += \text{client.weight}(\text{neuron}, \text{feature}) \times \text{feature.value}$\;
          }

          $\text{neuron2prov}[\text{neuron}][\text{client}] = \text{cont} \times \text{neuron2grad}[\text{neuron}]$\;
        }
      }

      \tcp{Section~\ref{sec:Measuring Client's Contribution}}
      $\text{client2contribution} = \{\}$\;
      \For{$\text{client}$ \textbf{in} $\text{clients}$}{
        \tcp{Equation~\ref{eq:client_k_total_contrib}}
        $\text{contribution} = 0$\;
        \For{$\text{neuron}$ \textbf{in} $\text{neuron2prov}$}{
          $\text{contribution} += \text{neuron2prov}[\text{neuron}][\text{client}]$\;
        }
        $\text{client2contribution}[\text{client}] = \text{contribution}$\;
      }

      \tcp{Equation~\ref{eq:client_k_total_contrib_softmax}}
      $\text{client2norm\_contribution} = \text{Softmax(client2contribution.values)}$\;
      % $\text{total} = \text{sum}(\text{client2contribution.values()})$\;
      % \For{$\text{client}$ \textbf{in} $\text{client2contribution}$}{ \tcp{Equation~\ref{eq:client_k_total_contrib_softmax}}
      %     $\text{client2norm\_contribution}[\text{client}] = \exp(\text{client2contribution}[\text{client}]) / \text{total}$\;
      % }
      \KwRet{$\text{client2norm\_contribution}$}\;

    }
    \caption{\tool's Approach}
  \label{algo:provfl}}

\end{algorithm}

\section{{Design of \tool}}
\label{sec:provfl-design}
\tool addresses the aforementioned challenges using  \textbf{{\em neuron provenance}}. At a high level, \tool dynamically tracks the lineage of the global model at the neuron level and identifies the most influential clients against a given prediction by the global model (\globalModel) on an input. Enabling provenance at the neuron level solves the complexities of different neural networks architectural design (\eg number of layers and parameters, different activation functions) and enables \tool to work in cross domains such as image and text classification tasks.

In essence, \tool first identifies the influential neurons by jointly analyzing neuron activations, the layers the neurons are in, and gradients with respect to individual neurons in the global model for a given test input. Next, \tool precisely quantifies the individual contribution of each corresponding neuron from every participating client to the neurons in the global model.
Finally, \tool computes the total contribution of each client to the global model. These steps collectively construct a comprehensive end-to-end provenance graph, which is used to debug the contributions of clients in the given prediction of the global model. Algorithm~\ref{algo:provfl} outlines the design of \tool.

\subsection{{Determining Influential Neurons}}
\label{sec:Measuring Neuron's Contribution}

\tool first aims to identify neurons that actively participate in an FL global model's prediction. Traditional data provenance approaches must trace the participation of all input data records in the operation for completeness, eventually mapping them to individual outputs of the operation. However, tracking the provenance of all neurons with equal importance is wasteful because not all neurons participate equally in a model's prediction. Therefore, tracking the behavior of neurons with the same importance in a model may lead to over-approximation (\ie more than expected clients are classified as contributors) when provenance is used to identify the contributing clients.

The behavior of a neural network on a given input is determined by the set of activated neurons in the network, and different sets of neurons are activated on different inputs. We leverage this insight and apply \tool's neuron provenance to dynamically quantify the influence of global model neurons on each prediction for the given input. This reduces the likelihood of over-approximation by minimizing the contribution of neurons that may distort the outcome when the lineage of a specific neuron is used to localize the influential client.

Mathematically, the output of a neuron is $z = \sigma (\mathbf{w} \cdot \mathbf{z} )$, where $\mathbf{w}$ is the set of weights of the neuron, $\mathbf{z}$ is the input to a neuron, and $\sigma$ is the activation function. One of the commonly used activation functions ($\sigma$) is ReLU ~\cite{10.5555/3104322.3104425}. The output of $\sigma$ is called the activation or output of the neuron. A neuron with ReLU function is considered active if $z > 0$. Note that the output of a neuron ($z$) is part of the input to the neurons of the next layer.  Next, \tool computes the activation of each neuron in the network. Suppose that $n_j$ represents the $j$-th neuron in a neural network and the set of all the outputs (\ie activations) of all the neurons in a neural network can be represented as $\{z_{n_1}, z_{n_2}, ..., z_{n_j}\}$, which captures the complete dynamic behavior of the network on a given test input $\mathbf{x}$. Note that for the first layer, the input $\mathbf{z}$ to neurons will be the input $\mathbf{x}$ to the model \ie $\mathbf{z} = \mathbf{x}$ for the first layer of the neural network.

After computing the global model neurons activations, \tool's goal is to find their measurable contribution towards the global model's prediction ($y$) on an input. In the output ($y$) of the global model, not all the neurons carry equal importance. For instance, neurons in the last layers learn better and more rigorous features than neurons in the initial layers of the network~\cite{olah2018building}. Since \tool aims to localize the client that contributed the most towards a prediction, assigning equal importance to all neurons will again cause over-approximation or even wrong client localization. To enable precise and accurate provenance, we must measure the individual influence of a neuron on the final prediction.

\tool quantifies the impact of the output of a neuron on the global model's prediction by computing the gradient \wrt every activated neuron on a given input to \globalModel. Similar to taint analysis in program analysis, gradients are sophisticated taints that encapsulate the impact of a neuron output on the output ($y$) of the global model. The intuition behind this is that the neurons with a higher gradient will likely cause a bigger change in prediction. Thus, such neurons are likely to be more influential to a model prediction. We use the aforementioned insight to find the influence of a neuron in the prediction ($y$) of the global model. The influence, denoted by $c_{n_j}$, of a neuron $n_j$ in the output ($y$) is the partial derivative of $y$ with respect to $z_{n_j}$, which measures how much $y$ changes when $z_{n_j}$ changes slightly. Mathematically, we write it as:

\vspace{-2ex}
\begin{equation} \label{eq:neuron-contribution}
  c_{n_j} = \frac{\partial y}{\partial z_{n_j}}
\end{equation}

\tool computes the gradients using the automatic differentiation engine of PyTorch~\cite{pytorch}. \tool starts from the output layer and goes back to the input layer, using the chain rule of differentiation at each step. By the end of this phase, \tool determines the gradient (influence) of global model neurons on its output ($y$). For instance, in the presence of a disease in a medical imaging input (\eg predicting colorectal cancer (CRC) from histological slides of tumor tissue), the fused neurons of the global model that have learned the representation of that particular disease during FL training will significantly influence the model's output ($y$). These gradients are essential in mapping neurons of clients' models to the most influential ones in the global model.

\subsection{{Neuron Provenance Across Fusion}}
\label{sec:Neuron-Provenance-Across-Fusion}
In this step, \tool accurately determines the individual contribution of each corresponding neuron from every participating client to the neurons of the global model. In essence, \tool maps the outputs of the global model neurons to clients' neurons during prediction. Finding such a mapping and its magnitude has two challenges. First, FL uses fusion algorithms to merge clients' neurons statically. Instrumenting the fusion algorithms to trace the flow of weights across fusion is prohibitively expensive, as numerous clients participate in a round where each model may have millions of neurons. Second, the influence of clients' neurons on the neurons of the global model (\globalModel) is directly impacted by the output of the preceding layer in the global model, \ie the output of the neuron in the global model's previous layer is the combined output of the corresponding neurons of each client in that layer.  Consequently, attempting to determine clients' neurons' contributions by feeding input to the clients' model in isolation will lead to incorrect neuron provenance, as it cannot capture the overall impact of other clients.

\tool leverages the insight that the set of weights of a single neuron in the global model is determined by the corresponding weights of the neurons in the clients' models. Mathematically, the weights of a single neuron in the global model, represented as $\mathbf{w_{g}} = [ w_{g}^{1}, w_{g}^2, \cdots, w_{g}^{i}] $, are given by the following equation:
\vspace{-2ex}
\begin{equation} \label{single_weight2}
  \begin{split}
    w_g^{i} & = \sum_{k=1}^{K} p_k * w_k^{i} \\
    & = p_1 * w_1^{i} + p_2 * w_2^{i} + \cdots + p_k * w_k^{i}
  \end{split}
\end{equation}

Here, $w_{k}^{i}$ is the $i$-th weight of the neuron in the $k$-th client model. The variable $p_k$ is $n_k/n$, where $n_k$ represents the size of training data of client $k$, and $n$ represents the total number of data points from all clients, and it is calculated as $n = \sum_{k=1}^{K} n_k$ (Equation~\ref{eq:1}). Given an input $\mathbf{z}$ to the neuron $\mathbf{w_{g}}$ of \globalModel,  a client's contribution can be calculated as follows:

\vspace{-2ex}
\begin{equation} \label{single_weight3}
  \begin{split}
    z_{out} & = \mathbf{w_{g}} * \mathbf{z} \\
    & = [w_{g}^1, w_{g}^2, \cdots, w_{g}^{i}] * [z^1, z^2, \cdots, z^i] \\
    & = w_{g}^1 * z^1 + w_{g}^2 * z^2 + \cdots + w_{g}^{i} * z^i \\
    & = [p_1*w_1^1 +  p_2*w_2^1 + \cdots + p_k*w_k^1] * z^1 \\
    & + [p_1*w_1^2 +  p_2*w_2^2 + \cdots + p_k*w_k^2] * z^2 \\
    & + \cdots \\
    & + [p_1*w_k^i +  p_2*w_2^i + \cdots + p_k*w_k^i] * z^i
  \end{split}
\end{equation}

Here, $z^i$ is the $i$-th input feature to the neuron and $z_{out}$ is the output of the neuron. Thus, the contribution of a client $k$, denoted by $[t_k]$, in a neuron $n_j$ of the global model (\globalModel) is given by the following equation:

\vspace{-2ex}
\begin{equation}\label{eq:client_k_neuron_contrib}
  \begin{split}
    [t_k]_{n_j}  & =  (p_k * w_k^{1} * z^1 + p_k * w_k^{2} * z^2 + \cdots \\
    & + p_k * w_k^{i} * z^i) * c_{n_j} \\
    & = (p_k * [w_k^{1} * z^1 + w_k^{2} * z^2 + \cdots + w_k^{i} * z^i]) * c_{n_j} \\
    & = c_{n_j} * p_k * \sum_{i=1} w_k^{i} * z^i
  \end{split}
\end{equation}
In the above equation, $p_k * \sum_{i=1} w_k^{i} * z^i$ is the exact contribution of a client $k$ in a neuron $n_j$ of the global model. The global gradient of neuron $n_j$ is $c_{n_j}$ which is multiplied with client contribution to find its actual contribution (\ie influence) towards the prediction of the global model. For instance, if the contribution of a client $k$ is high in a neuron $n_j$ but globally the neuron $n_j$ has minimal influence on the global model prediction then $c_{n_j}$ will scale down the contribution of the client in the given neuron $n_j$. Note that $z^i$ represents the $i$-th output of the previous layer in the global model during prediction. At the end of this stage, \tool constructs a \textbf{neuron provenance} graph that traces a global model's prediction to influential neurons in the global model (\globalModel), which are further traced back to individual neurons in the clients' models.

\subsection{{Measuring Client's Contribution}}
\label{sec:Measuring Client's Contribution}
To find the end-to-end contribution, we must accumulate neuron-level provenance, $c_{n_j} * p_k * \sum_{i=1} w_k^{i} * z^i$,  of a given client's model to derive its complete contributions toward the global model's prediction. A client's overall contribution to the global model prediction is determined by the sum of the client's contribution to the neurons of the global model. Specifically, if the set of neurons of the global model is denoted by $n_1, n_2, \cdots, n_j$, then the total contribution ($T_k$) of the client $k$ can be calculated using Equation~\ref{eq:client_k_neuron_contrib} as follows:

\vspace{-3ex}
\begin{equation}\label{eq:client_k_total_contrib}
  \begin{split}
    T_k & = \beta_{n_1} * [t_{k}]_{n_1} + \beta_{n_2} * [t_{k}]_{n_2} + \cdots + \beta_{n_j} * [t_{k}]_{n_j} \\
    & = ([c_{n_1} * \sum_{i=1} w_{k\_n_1}^{i} * z_{n_1}^i]_{n_1} + [c_{n_2} * \sum_{i=1} w_{k\_n_2}^{i} * \\
    & z_{n_2}^i]_{n_2} + \cdots + [c_{n_j} *\sum_{i=1} w_{k\_n_j}^{i} * z_{n_j}^i]_{n_j}) * p_k
  \end{split}
\end{equation}

$\beta$ is an importance factor that \tool computes using an exponential decay method for each neuron based on its position in the neural network. Specifically, \tool assigns higher importance to the last layers and lower importance to the earlier layers to minimize the noisy contributions, based on the evidence presented elsewhere ~\cite{olah2018building}. $[t_k]_{n_j}$ is the contribution of the client $k$ in neuron $n_j$, $z_{n_j}^i$ is the $i$-th input feature to neuron $n_j$, and $w_{k\_n_j}^{i}$ is the $i$-th weight of neuron $n_j$ in the client $k$ model. Using Equation~\ref{eq:client_k_total_contrib} we can compute, for each client $k$, the total contribution towards the global model prediction. Thus, the client with max contribution is the client that has the most influence on the global model prediction. To make the client contribution more interpretable, we normalize the client contribution by using the softmax function as follows:

\vspace{-3ex}
\begin{equation}\label{eq:client_k_total_contrib_softmax}
  \begin{split}
    \tilde{T_k} & = \frac{e^{T_k}}{\sum_{i=1}^{K} e^{T_i}}
  \end{split}
\end{equation}

$\tilde{T_k}$ is the normalized contribution of the $k$-th client, which is now a probability value between 0 and 1, representing the relative influence of client-$k$ on the global model output $y$ for a given input.

\tool concludes its \textbf{neuron provenance} capturing technique by listing the total contribution of each participating client in an FL round towards a global model's prediction on a given input. The magnitude of the contributions can be interpreted as a confidence level of \tool in identifying the source of the global model's prediction. Given that the total confidence scores of all clients cannot exceed 1, if a client has a contribution score of 0.6, it implies that no other client can surpass a score of 0.4. This makes the client most influential in determining the global model prediction and most likely responsible for the prediction.

\noindent{\textit{\textbf{Enable \tool to Use GPU.}}}  By design, \tool is compatible with hardware accelerators and can fully harness their parallelizability. The primary dependency of \tool is capturing the output of previous layer neurons in the global model for input to the next layer neurons, which inherently exists in inference as well. Additionally, \tool computes gradients using Equation~\ref{eq:neuron-contribution}, leveraging the chain rule of differentiation that the hardware accelerators can parallelize.
Next, Equation~\ref{eq:client_k_neuron_contrib} dissects the global model neuron and computes the contribution based on the previous layer neurons' outputs ($\mathbf{z}$) of the global model, which is the cumulative output of all clients' neurons in that previous layer. This is the only dependency in \tool.
Once \tool has the cumulative output from the previous layer neurons, it parallelizes the process to find the contribution of a client in each neuron of the global model in the current neuron layer and ultimately the total contribution using Equations~\ref{eq:client_k_total_contrib} and ~\ref{eq:client_k_total_contrib_softmax}. These optimizations in \tool enable neuron-level provenance for neural networks primarily deployed on GPUs, such as GPT.

\section{Experimental Evaluations}

We design experiments to evaluate \tool's accuracy in localizing the client responsible for a global model's prediction on an input. We ask the following research questions.
\begin{itemize}
  \item How accurate is \tool in identifying the client(s) responsible for a global model's prediction?
  \item  Is \tool equally accurate on FL of different models and architectures such as CNNs and transformers (GPT)?
  \item How accurate \tool is in localizing clients responsible for mispredictions by global model?
  \item Does \tool remain effective with varying data distributions and differential privacy?
  \item Can \tool scale to a large number of clients?
    %    \item Can \tool work when differential privacy is enabled in FL?
  \item What is the runtime performance of \tool?
\end{itemize}

\noindent\textbf{\textit{Models and Datasets.}} We evaluate \tool on state-of-the-art and commercially used CNNs, including ResNet-18~\cite{he2015deep} and DenseNet-121~\cite{huang2017densely}, as well as the two most popular transformer models, BERT~\cite{Devlin2019BERTPO} and GPT~\cite{radford2018improving} to demonstrate the wide applicability of \tool.  We train ResNet and DenseNet on CIFAR-10~\cite{cifar10} and MNIST~\cite{lecun-mnisthandwrittendigit-2010}. These network-dataset combinations are widely used and serve as standardized benchmarks in practice~\cite{mcmahan2017communication,li2022federated}. We also evaluate \tool on real-world medical imaging datasets, including the Colon Pathology dataset~\cite{kather2019predicting} and Abdominal CT dataset~\cite{xu2019efficient, bilic2023liver}, to demonstrate its usability in complex real-world FL systems. The Colon Pathology dataset contains 107,180 biomedical images representing nine classes of colon pathology, while the Abdominal CT dataset contains 58,830 images of abdominal CT scans representing 11 classes. More details about these datasets can be found in~\cite{medmnistv1, medmnistv2}. For NLP tasks, we evaluate \tool on BERT and GPT models trained on the DBpedia and Yahoo Answers datasets~\cite{zhang2015character}. The DBpedia dataset contains 560,000 training samples and 70,000 testing samples, while the Yahoo Answers dataset contains 1,400,000 training samples and 60,000 testing samples, representing 14 and 10 classes, respectively.

\noindent\textbf{\textit{Data Distribution Among Clients}}. We use Dirichlet distribution in FL to distribute non-overlapping data points among clients in each round. This is the standard FL data distribution method proven to produce real-world distribution~\cite{Wang2020Federated, wang2020tackling, lin2020ensemble, li2022federated}. The parameter ($\alpha$) in Dirichlet ranges from [0, $\infty$), determining the level of Non-IID  in experiments. For instance, when $\alpha$ equals 100, it replicates uniform local data distributions, while smaller $\alpha$ values increase the probability that clients possess samples from a single class~\cite{lin2020ensemble}.
A value of 0.5 is a common practice in prior work~\cite{Wang2020Federated, li2022federated}. We use an even stricter parameter value of 0.3 to stress test \tool and demonstrate its usability in more challenging cases. Nevertheless, Section~\ref{sec:data-distribution} performs sensitivity analysis by varying $\alpha$ from 0.1 to 1. These settings inherently simulate varying degrees of label overlap among clients. To explicitly manage overlapping labels, pathological data distributions can be employed~\cite{li2022federated}, as shown in Figure~\ref{fig:motivating-example}. The pathological data distribution is available in \tool's artifact.
Furthermore, \tool's artifact contains configurable data distributions among clients and allows evaluations on varying numbers of test inputs.

% \tool generates a ranked list of clients according to their responsibility for a given prediction. 

% For instance, Figure~\ref{fig:motivating-example} demonstrates how differences in the data distribution of the same label (e.g., ``Mucus") influence client rankings (H2, H4, and H3). Even when each client contributes to multiple labels, \tool distinguishes clients based on how their unique data influences model updates.  }  

\noindent\textbf{\textit{Experimental Environment.}} To resemble real-world FL, we deploy our experiments in Flower FL~\cite{beutel2020flower}, running on an enterprise-level cluster of six NVIDIA DGX A100~\cite{NVIDIADG22:online} nodes. Each node is equipped with 2048 GB of memory, at least 128 cores, and an A100 GPU with 80 GB of memory.

We vary training rounds between 15 to 80  with clients ranging from 100 to 1000, thus testing \tool on more configurations than any related work~\cite{lin2020ensemble, liu2021projected}. Ten randomly selected clients participate in each round, reflecting a real-world scenario where not all the clients participate in the given round~\cite{Li2020Fair}. We evaluate \tool with FedAvg~\cite{mcmahan2017communication}.

\noindent\textbf{\textit{Localization Accuracy.}} To measure the performance of \tool, we evaluate the accuracy of \tool in finding the responsible clients. For brevity, we refer to this as localization accuracy, which is defined as follows: Given the $z$ number of test inputs to the global model (\globalModel), if \tool accurately locates $m$ times the clients responsible for the $z$ predictions, then the localization accuracy is $\frac{m*100}{z}$.

\begin{figure}[t]
  \centering

  \includegraphics[width=0.49\textwidth]{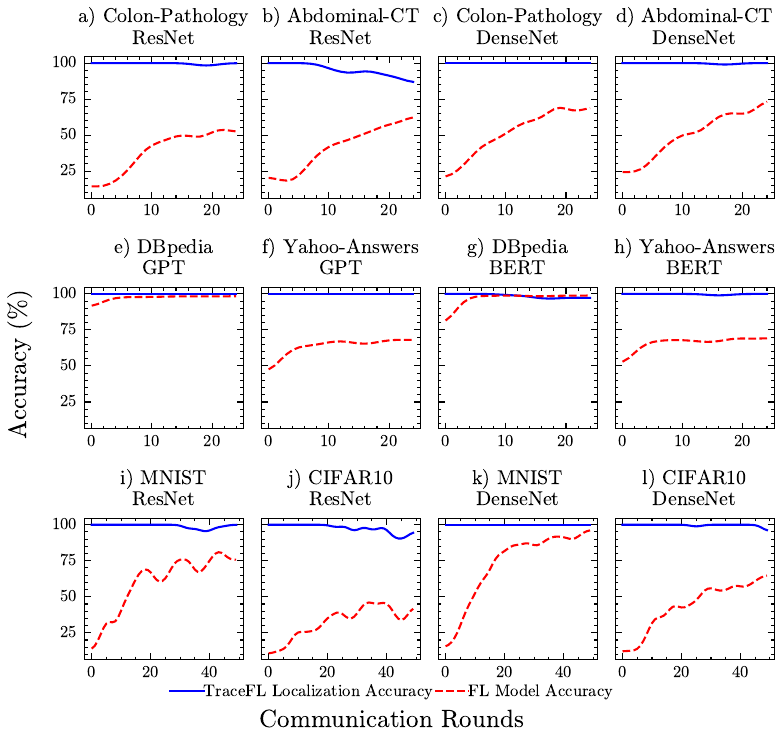}

  \caption{\tool performance on multiple datasets and models both on text and image classification tasks.}
  \label{fig:tracefl-main-result}
  \vspace{-2ex}
\end{figure}

\subsection{{\tool's Localization Accuracy in Correct Predictions}}
\label{sec:correct-predictions-localization}

Identifying the clients most responsible for correct prediction is a key debugging objective that helps encourage future participation of those clients to improve the overall FL accuracy. Note that \tool directly does not improve the FL model accuracy. Instead, it reasons  about the behavior of the FL global model which an FL developer can use to improve the FL model accuracy (\eg selecting clients which are contributing more in the FL global model predictions).

\tool's \textbf{neuron provenance} traces predictions back to clients trained on those labels, ranking clients by their contribution. \tool returns a ranked list of clients in descending order of responsibility towards a prediction, where the client with the highest score is likely to be most responsible.

We evaluate \tool's localization accuracy on two real-world medical imaging datasets, two standardized image datasets, and two NLP classification datasets using ResNet, DenseNet, BERT, and GPT models resulting in over 12 FL configurations spanning a total 400 FL rounds and 4000 models.
We verify if the most responsible client returned by \tool contains the data with the label that was correctly predicted by the global model. We measure the accuracy on at least 10 test inputs in each round. Figure~\ref{fig:tracefl-main-result} shows \tool's performance in localizing responsible clients.
The X-axis represents training rounds, while the Y-axis shows the FL global model's classification accuracy and \tool's localization accuracy.

We include the FL global model's accuracy to demonstrate the training progression. Higher global model accuracy improves neuron provenance confidence, aiding \tool's effectiveness. Global model accuracy helps calibrate the provenance results because lower model accuracy leads to low confidence in prediction, which transitively reduces the confidence of neuron provenance, causing additional challenges for \tool. As training progresses and more clients with unique labels participate, the global model's accuracy improves.

Our results indicate that \tool consistently localizes responsible clients regardless of the global model's performance, neural network architecture, number of training rounds, or dataset. It accurately identifies contributions even from clients participating for the first time. Across different FL settings, \tool's average localization accuracy on image classification tasks is 98.96\%, and in text classification tasks, it is 99.59\%, demonstrating its broader effectiveness and applicability to domains other than image classification.

\noindent{\textbf{\textit{\underline{Takeaway.}}}} On average, \tool achieves localization accuracy of 99.12\% across all FL experiments settings.

\subsection{{\tool's Localization Accuracy in Mispredictions}}
\label{sec:faulty-clients}

\begin{table}[t]
  \centering
  % \begin{tabularx}{0.4\textwidth}{|X|X|X|}

\begin{tabularx}{0.48\textwidth}{|l|l|X|X|X|}
\hline
\textbf{Domain} & \textbf{Dataset} & \textbf{Dirichlet Distribution ($\alpha$)} & \textbf{FedDebug Accuracy (\%)} & \textbf{TraceFL Accuracy (\%)} \\
\hline
\multirow{12}{*}{Image} & \multirow{3}{*}{Abdominal-CT} & 0.3 & 0.00 & 100 \\
& & 0.7 & 21.5 & 100 \\
& & 1.0 & 44.4 & 100 \\
\cline{2-5}
& \multirow{3}{*}{Colon-Pathology} & 0.3 & 0.00 & 100 \\
& & 0.7 & 54.7 & 100 \\
& & 1.0 & 68.7 & 100 \\
\cline{2-5}
& \multirow{3}{*}{CIFAR10} & 0.3 & 20.0 & 100 \\
& & 0.7 & 11.3 & 100 \\
& & 1.0 & 22.0 & 100 \\
\cline{2-5}
& \multirow{3}{*}{MNIST} & 0.3& 14.0 & 100 \\
& & 0.7 & 86.0 & 100 \\
& & 1.0 & 36.0 & 100 \\
\hline
\multirow{6}{*}{Text} & \multirow{3}{*}{DBpedia} & 0.3 & NA & 96.7 \\
& & 0.7 & NA & 94.0 \\
& & 1.0 & NA & 97.3 \\
\cline{2-5}
& \multirow{3}{*}{Yahoo-Answers} & 0.3 & NA & 100 \\
& & 0.7 & NA & 100 \\
& & 1.0 & NA & 100 \\
\hline
\end{tabularx}

  \caption{Comparison of \tool with FedDebug on localizing clients responsible for misprediction. FedDebug is compatible with image classification only and is effective under specific data distribution (\ie$\alpha=1$).}
  \label{table:tracefl_and_feddebug_results}
  \vspace{-3ex}
\end{table}

FL's global model can exhibit unwanted behavior (\eg mispredictions) due to intentional or unintentional faults in the training data of clients. Mislabelling in training data may occur due to faulty sensors, human error in labeling data, or, in some cases,  adversarial attacks~\cite{mcmahan2017communication, jiang2020federated, rieke2020future, long2020federated, zheng2021applications}.
Finding a client responsible for such behavior is a crucial debugging goal that helps FL developers exclude such clients from participating in future rounds to improve the global model's quality.

To evaluate \tool's localization accuracy on mispredicted labels by a global model, we design the following experiments with ten clients. Similarly to prior work on fault localization in FL, FedDebug~\cite{feddebug}, we select one client in an FL round and flip a specific label in its training data to make it faulty. The inclusion of such clients influences the global model to make mispredictions. For instance, in the medical dataset, we flip the label `Cancer-associated Stroma' to  `Adipose' in the Colon Pathology dataset to reflect a faulty hospital containing incorrect label data that may occur due to misdiagnosis.

Table~\ref{table:tracefl_and_feddebug_results} shows the results. \tool outperforms FedDebug significantly and can operate in cross-domain tasks of image and text classification without any change in its approach. This is expected since FedDebug, by construction, applies to a different problem setting, i.e., debugging the model instead of the prediction, and it primarily targets a specific set of Non-IID data distributions. 
Even on image classification tasks, \tool outperforms FedDebug in terms of localization accuracy. For instance, in Abdominal-CT with $\alpha=1$, FedDebug's average accuracy is 44.4\% while \tool's accuracy is 100\%.

\noindent{\textbf{\textit{\underline{Takeaway.}}}} \tool achieves 99.3\% average localization accuracy across 18 FL settings, whereas FedDebug's average localization accuracy is only 32\% on image classification.

\subsection{{ \tool's Robustness}}
Varying the client's data distribution and applying differential privacy (DP) techniques in FL pose additional hurdles to FL in achieving high model accuracy, which, in turn, may pose challenges to \tool in keeping its high localization accuracy. Therefore, to add rigor to our experiments, we evaluate the impact of these two additional FL settings on \tool localization accuracy. In this section, we only include results from the most challenging experiment setting due to space constraints.

\begin{figure}[t]

  \includegraphics[width=0.49\textwidth]{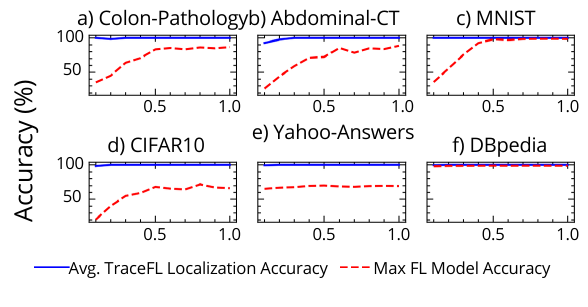}

  \caption{\tool performance on different data distributions. The X-axis represents the values of Dirichlet alpha.
    %As the value of $\alpha$ decreases, the FL training accuracy decreases. \tool's performance remains consistent across different data distributions.
  }
  \label{fig:dirichlet_distribution_results}
  \vspace{-3ex}
\end{figure}

\subsubsection{\textbf{Varying Data Distribution}}
\label{sec:data-distribution}
Different distributions of data among clients can impact the FL training process. For instance, in a highly challenging data distribution ($\alpha = 0.1$), FL training suffers from low global model accuracy.
This is a known phenomenon in FL~\cite{li2022federated}, where the FL fusion algorithm struggles to aggregate clients' models trained on severely heterogeneous training data. To mitigate bias towards a specific Dirichlet data distribution, we evaluate \tool on varying the value of $\alpha$ from 0.1 to 1.0, showing the impact of different data distributions on \tool's localization accuracy.

Figure~\ref{fig:dirichlet_distribution_results} shows the results of this experiment on all six datasets. The X-axis represents the value of $\alpha$ in the Dirichlet distribution, while the Y-axis represents the accuracy. For a value of $\alpha$, we report the maximum accuracy achieved by the global model across all the rounds as FL model accuracy and the average accuracy of \tool across all the rounds as localization accuracy of \tool.

As expected, the FL training accuracy decreases as the value of $\alpha$ decreases. This is because the clients have varying data both in terms of quantity and labels. For instance, when $\alpha = 0.1$ in Figure~\ref{fig:dirichlet_distribution_results}-(a), the maximum FL global model accuracy observed across all rounds is 35.25\% and when $\alpha = 0.5$ the maximum accuracy is 83.4\%. Since GPT is an advanced neural network architecture that learns better in comparison to DenseNet, the FL training accuracy is higher in GPT on lower $\alpha$ values as well. Overall, \tool localization accuracy is 99.76\%, on average, across all values of $\alpha$. The line plots show no significant change in \tool localization accuracy, demonstrating \tool's robustness in challenging real-world data distributions.

\subsubsection{\textbf{Differential Privacy-Enabled FL}}
\begin{table}[t]
  \centering
  % \begin{tabularx}{0.4\textwidth}{|X|X|X|}

\begin{tabularx}{0.48\textwidth}{|X|X|X|X|}  
\hline
\textbf{DP Noise} & \textbf{DP Sensitivity}  & \textbf{FL Model Accuracy \%} & \textbf{\tool Avg. Accuracy \%} \\
\hline
0.003 & 15 & 97.36 & 100 \\
\hline
0.006 & 10 & 97.90 & 100 \\
\hline
0.012 & 15 & 88.81 & 100 \\
\hline\end{tabularx}

  \caption{Results of \tool with DP in FL.}
  \label{table:differential_privacy_results_table}
  \vspace{-2ex}
\end{table}

\begin{figure}[t]
  \centering

  \includegraphics[width=0.49\textwidth]{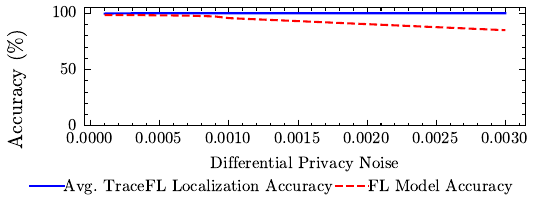}

  \caption{Impact of DP noise on FL training accuracy.
    %DP sensitivity is 50. As the noise increases, the FL training accuracy decreases. However, \tool's performance remains consistent.
  }
  \label{fig:differential_privacy_noise_impact}
\end{figure}

Differential privacy (DP) is a privacy-preserving mechanism that ensures that the output of a model does not reveal any information about the individual data points. DP in FL~\cite{mcmahan2018learning} adds noise to the weights of a model to protect against an adversary stealing or recovering the individual training data points. However, a delicate balance is needed in DP between the noise to be added and model accuracy, as adding too much noise severely decreases the model's accuracy.

We evaluate \tool's robustness when DP is enabled in FL, using standard DP settings in FL that provide optimal privacy and model accuracy, as mentioned in prior work~\cite{mcmahan2018learning}. Table~\ref{table:differential_privacy_results_table} presents the results of this experiment, and Figure~\ref{fig:differential_privacy_noise_impact} shows the impact of noise on the FL training accuracy. As expected, the FL model's accuracy decreases when the DP noise increases and vice versa.

However, \tool maintains its performance in DP-enabled FL. As DP adds noise to the model weights, the global model's output is still based on its neurons' activations on the given input. Thus, \tool's working principle remains intact, and it successfully traces back to the source of the prediction based on the global model's \textbf{neuron provenance}. We want to emphasize that \emph{\tool does not recover the individual clients' data points. It only identifies the responsible clients in ranked order.}
Overall, we find that \tool is robust against the use of differential privacy in FL where it achieves an average localization accuracy of 99\% in GPT and DBpedia dataset (Figure~\ref{fig:differential_privacy_noise_impact} and Table~\ref{table:differential_privacy_results_table}).

\noindent{\textbf{\textit{\underline{Takeaway.}}}}
\tool is robust to challenging real-world data distributions and the use of differential privacy,  achieving approximately 99\% localization accuracy.

\subsection{\tool's Scalability}

\begin{table}[t]
  \centering
  \begin{tabularx}{0.48\textwidth}{|X|X|X|}
\hline
\textbf{Total Clients}  & \textbf{FL Model Accuracy \%}   & \textbf{\tool Avg. Accuracy \%}  \\
\hline
200 & 98.49 & 99.76 \\\hline
400 & 98.29 & 99.76 \\\hline
600 & 98.39 & 100 \\\hline
800 & 98.10 & 100 \\\hline
1000 & 98.05 & 99.52 \\\hline

\end{tabularx}

  \caption{Scalability results of \tool with different number of clients with GPT.}
  \label{table:scalability_results_table_total_clients}
  \vspace{-3ex}
\end{table}

\begin{figure}[t]
  \centering

  \includegraphics[width=0.49\textwidth]{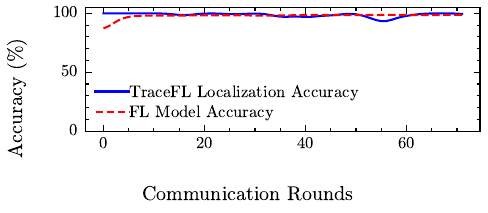}

  \caption{\tool's scalability when \# of rounds increase}
  \label{fig:scalability_results_400_clients_rounds_80}
  \vspace{-3ex}
\end{figure}

We assess the scalability of \tool across three different dimensions: (1) by increase the total clients, (2) by increasing the client participation, and (2) by increasing the number of rounds.  First, we vary the number of clients from 200 to 1000 and measure if \tool can still accurately localize the responsible client. We use the state-of-the-art neural network GPT and the DBpedia dataset.
Table~\ref{table:scalability_results_table_total_clients} presents the results of the scalability experiment. We observe that \tool's performance remains consistent, with an average localization accuracy of 99\% across 200 to 1000 clients over a total of 75 FL training rounds. This experiment significantly exceeds the scale of experiments performed by prior work~\cite{feddebug}.

When we vary the number of participating clients per round from 20 to 50, \tool's performance remains stable, achieving 100\% localization accuracy across 60 FL training rounds. Prior work has shown that even at an enterprise scale, only a few clients participate in a single FL round~\cite{bonawitz2019towards, konečný2018federated}. Furthermore, we evaluate the scalability of \tool over up to 80 rounds with 400 clients in total. Figure~\ref{fig:scalability_results_400_clients_rounds_80} demonstrates that \tool maintains consistent performance with an average localization accuracy of 98\%. These results indicate that \tool is scalable and can handle numerous clients and rounds without compromising its performance.

\noindent{\textbf{\textit{\underline{Takeaway.}}}}
Overall, \tool is capable of handling the provenance of millions of neurons in the neural network to accurately identify the most responsible client. In FL settings of up to 80 FL training rounds and 1000 clients using large models such as GPT and BERT—and 6 different datasets, \tool achieves an average localization accuracy of 99.20\%.

\subsection{\tool's Localization Time}
We evaluate the runtime performance of \tool by measuring the time \tool takes to accurately localize the responsible clients in FL. As mentioned before, there is no existing method that localizes the responsible clients for both correct and incorrect predictions. The closed related work to \tool is FedDebug, which only localizes faulty clients. Thus, we compare \tool's localization time with FedDebug's faulty localization time.

\begin{figure}[t]
  \centering
  \includegraphics[width=0.49\textwidth]{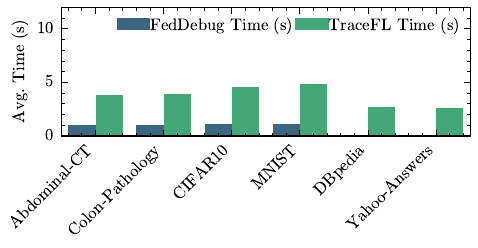}

  \caption{Client localization of \tool vs. FedDebug.}
  \label{fig:overhead-comparison}
  \vspace{-2ex}
\end{figure}

Figure~\ref{fig:overhead-comparison} presents the localization times per dataset for both \tool and FedDebug, averaged across faulty client localization settings. Note that FedDebug is not compatible with the text classification models; therefore, its localization times for the two text datasets are not available. \tool takes, on average, 3.7 seconds to localize the responsible client, whereas FedDebug's faulty client's localization time is 1.1 seconds on average. This is expected as \tool requires computing gradients of neuron outputs, whereas FedDebug compares raw neuron activations.  While \tool's localization time is higher than FedDebug, it is almost negligible compared to the FL's per round training time (in minutes if not hours~\cite{feddebug}).

\noindent{\textbf{\textit{\underline{Takeaway.}}}}
\tool compensates for the marginally slower localization time with much broader debugging support for model architecture, text data domains, and general-purpose reasoning in FL.

\subsection{Threat to Validity and Limitations}
There are two primary threats to the validity of the results. First, in our experiments, we select a random subset of clients to participate in every round. A different sequence of randomly selected participating clients may alter the \tool's accuracy. We mitigate this threat by performing responsible client localization on every round and then reporting the average localization accuracy across rounds. Second, the same Dirichlet distribution ($\alpha$) may provide a different distribution of the training data across clients. Even when the value $\alpha$ is the same, the localization accuracy of \tool may vary slightly. We mitigate this threat by averaging the localization accuracy across rounds and also measuring the localization accuracy on different datasets and models.  \tool is designed for classification tasks in FL and may not be directly applicable to non-classification tasks such as text generation~\cite{sani2024future} and embeddings generation~\cite{gill2024meancacheusercentricsemanticcache}.

\section{Related Work}

\noindent{\textit{\textbf{Debugging and Interpretability in Machine Learning.}}} As the complexity of neural network models continues to increase, the need for interpretability techniques becomes more crucial and important. Interpretability techniques are used to understand the inner workings of a neural network. These techniques try to explain the decisions made by the model, and how the model makes these decisions. This is important for many reasons, including the ability to explain which input features are important to a model's output, to understand the model's behavior, and to identify potential biases and errors in a trained model. Several approaches, such as Integrated Gradients~\cite{sundararajan2017axiomatic}, Gradient SHAP~\cite{lundberg2017unified}, DeepLIFT~\cite{shrikumar2017learning}, Saliency~\cite{DBLP:journals/corr/SimonyanVZ13}, Guided GradCAM~\cite{selvaraju2017grad}, Occlusion (also called sliding window method)~\cite{zeiler2014visualizing}, and LIME~\cite{ribeiro2016should}, exist which evaluate the contribution of each input feature to model's output. For instance, Integrated Gradients~\cite{sundararajan2017axiomatic} evaluates the contribution of each input feature by calculating the integral of gradients \wrt input.  This is done along the path from a selected baseline to the given input. Occlusion involves replacing each contiguous rectangular region with a predetermined baseline or reference point and measuring the difference in the model's output. This approach is based on perturbations and provides a way to evaluate the importance of input features by measuring the change in the model's output.

Existing debugging techniques \cite{sun2022causality, usman2021nn, gerasimou2020importance, 10.1145/3490489, tao2023dlregion} are designed to identify issues and enhance the performance of a single neural network in centralized ML. These methods typically require access to training data, which is prohibited in FL. For example, NPC~\cite{10.1145/3490489} constructs a Decision Graph using training data. Furthermore, these approaches have not been evaluated on modern neural network architectures such as Transformers.

Almost all existing debugging and interpretability approaches are inapplicable in FL, as by design, they solve an orthogonal problem --- identifying the important feature in the input responsible for a prediction instead of clients. This distinction is critical because the training data or the training process is completely inaccessible in FL. Existing approaches require access to the client's data. Furthermore, they are only designed for a single neural network, but the FL global model is a mixture of clients' models participating in the given round. Operating these techniques on FL would require us to first identify a suspicious client's mode--a problem that \tool solves. Even if such techniques are applied to a client's model, the resulting feedback is not immediately actionable and constructive. \tool is designed to address the limitations of the existing debugging approaches and added challenges of FL, such as distributed training, inaccessibility to clients, and the mixture of models.

FedDebug~\cite{feddebug} introduces differential testing in FL to identify faulty clients by capturing each client's activations for a given input and localizing the client(s) whose behavior deviates from others. Building on FedDebug, a backdoor detection technique in FL is presented in~\cite{gill2023feddefender}. Additionally, FedGT~\cite{xhemrishi2023fedgt} aims to identify malicious clients in FL; however, it is limited to scaling up to 15-30 clients and has not been tested on advanced architectures like GPT.

Despite their contributions, these existing methods target a narrower problem under a specialized setting. FedDebug is limited to image classification tasks using CNNs, restricting its applicability to Transformer architectures. Additionally, it is designed primarily for faulty clients and IID distributions~\cite{li2022federated}, as demonstrated in Table~\ref{table:tracefl_and_feddebug_results}. In contrast, \tool targets a broader debugging problem using a domain-agnostic and highly accurate client localization mechanism applicable to diverse neural network architectures, data types, and distributions through its novel fine-grained \textbf{neuron provenance}.

There has been recent work on ensuring accountability in FL systems. A vast majority of solutions leverage the blockchain to ensure accountability~\cite{flchain,zhang,kang,blockflow,blockfla}. Some of these works (BlockFLow~\cite{blockflow}, BlockFLA~\cite{blockfla}) design an FL system that uses the Ethereum blockchain to provide accountability and monetary rewards for good client behavior. However, all these systems require utilizing the blockchain and entail significant modifications to the existing FL system, presenting a barrier to adoption. \tool, in contrast, can work with any existing system without modifications.

\noindent{\textit{\textbf{\textbf{Provenance Approaches in ML.}}}  Provenance has been extensively studied for both ML and dataflow programs~\cite{10.14778/2095686.2095693,10.14778/3402755.3402768,akoush2013hadoopprov,10.1145/2523616.2523619,interlandi2015titian}. They address various issues such as reproducibility~\cite{DBLP:journals/corr/abs-2006-07484, souza2019provenance, samuel2021machine,  xin2021production, rupprecht2020improving}, provide debugging and testing granularities~\cite{interlandi2015titian}, explainability~\cite{akoush2013hadoopprov}, and mitigating data poisoning attacks~\cite{stokes2021preventing, 8473440, baracaldo2017mitigating}. In the context of machine learning, provenance tracks the history of datasets, models, and experiments. This information is used to select the interpretability of neural network predictions and reproducibility.  Provenance-based approaches are important to create ML systems that generate reproducible results \cite{DBLP:journals/corr/abs-2006-07484, souza2019provenance, samuel2021machine,  xin2021production, rupprecht2020improving}.  For instance, Ursprung~\cite{rupprecht2020improving} captures provenance and lineage by integrating with the execution environment and records information from both system and application sources of an ML pipeline. Ursprung does not require changes to the code and only adds a small overhead of up to 4\%.

\section{Conclusion}
We introduce the concept of neuron provenance and developed a debugging and interpretability tool, \tool, for FL. \tool accurately identifies the primary contributors to a global model's behavior. Our evaluations show that \tool achieves an impressive average localization accuracy of 99\%. Furthermore, \tool also outperforms the existing fault localization technique. We provide a reusable functional artifact of \tool in the Flower framework to have an immediate practical impact in real-world FL deployment, addressing the open challenges of debugging and interpretability in FL.

\section*{Acknowledgement}  We thank anonymous reviewers for providing valuable and
constructive feedback to help improve the quality of this work.
This work was supported in part by Amazon - Virginia Tech Initiative in Efficient and Robust Machine Learning, 4-VA, and the National Science Foundation award 2106420. We also thank the Advanced Research Computing Center at Virginia Tech and the Flower FL framework for their support in building and evaluating this work.

\bibliography{main}

\bibliographystyle{IEEEtran}

\end{document}